%% file: main.tex
\documentclass[10pt,twocolumn,letterpaper]{article}

\usepackage{iccv}              


\definecolor{iccvblue}{rgb}{0.21,0.49,0.74}
\usepackage[pagebackref,breaklinks,colorlinks,allcolors=iccvblue]{hyperref}

\usepackage{adjustbox}
\usepackage{multirow}

\def\paperID{5174} 
\def\confName{ICCV}
\def\confYear{2025}

\title{Causality-guided Prompt Learning for Vision-language Models via Visual Granulation}

\author{
Mengyu Gao\textsuperscript{1}\textsuperscript{2}
\hspace{0.5cm}
Qiulei Dong\textsuperscript{1}\textsuperscript{2}~\thanks{Corresponding author} 
\hspace{0.5cm}
\\
\textsuperscript{1} State Key Laboratory of Multimodal Artificial Intelligence Systems, \\
Institute of Automation, Chinese Academy of Sciences \\
\textsuperscript{2} School of Artificial Intelligence, University of Chinese Academy of Sciences \\
{\tt\small gaomengyu2021@ia.ac.cn}
\hspace{0.5cm}
{\tt\small qldong@nlpr.ia.ac.cn.}
}

\begin{document}
\maketitle
\input{sec/0_abstract}    
\input{sec/1_intro}
\input{sec/2_relate}
\input{sec/3_method}
\input{sec/4_experiment}
\input{sec/5_conclusion}
{
    \small
    \bibliographystyle{ieeenat_fullname}
    \bibliography{ICCV2025-Author-Kit-Feb/ref}
}

\end{document}

%% file: sec/0_abstract.tex
\begin{abstract}
Prompt learning has recently attracted much attention for adapting pre-trained vision-language models ({\it e.g.}, CLIP) to downstream recognition tasks. However, most of the existing CLIP-based prompt learning methods only show a limited ability for handling fine-grained datasets. To address this issue, we propose a causality-guided text prompt learning method via visual granulation for CLIP, called CaPL, where the explored visual granulation technique could construct sets of visual granules for the text prompt to capture subtle discrepancies among different fine-grained classes through casual inference. The CaPL method contains the following two modules: (1) An attribute disentanglement module is proposed to decompose visual features into non-individualized attributes (shared by some classes) and individualized attributes (specific to single classes) using a Brownian Bridge Diffusion Model; (2) A granule learning module is proposed to construct visual granules by integrating the aforementioned attributes for recognition under two causal inference strategies. Thanks to the learned visual granules, more discriminative text prompt is expected to be learned. Extensive experimental results on 15 datasets demonstrate that our CaPL method significantly outperforms the state-of-the-art prompt learning methods, especially on fine-grained datasets.

\end{abstract}

%% file: sec/1_intro.tex
\section{Introduction}
\label{intro}

\par CLIP\cite{CLIP}, a typical vision-language model pre-trained on large-scale image-text pairs, has shown its zero-shot recognition ability in many existing works\cite{HierarchyCLIP,CuPL,WaffleCLIP,Frolic}. Recently, prompt learning methods have attracted growing interests for further enhancing CLIP's performance on downstream recognition tasks\cite{ProMetaR,TransAgent,CoOp,PromptKD}.

\par In general, the existing prompt learning methods refine the inputs of CLIP\cite{CLIP} with learnable prompts\cite{QCoOp,COMMA,LoGoPrompt,AAPL}, so that the visual and textual features outputted from CLIP are aligned more effectively. The early global prompt learning methods\cite{KIM,KgCoOp,TCP,ProGrad} process the features holistically during prompt learning. Despite achieving improvements on coarse-grained recognition, they fall short in fine-grained recognition, since the subtle discrepancies among fine-grained classes cannot be captured by global features as indicated in \cite{AAPE,CPL}. Recently, local prompt learning methods are proposed to focus on specific attributes, either identifying the most discriminative attributes while discarding others\cite{DEAL,GalLop,HPT}, or treating all attributes equally to encode attribute-specific representations\cite{CoCoLe,CPL,CDC}. However, these methods cannot allocate dynamic attention to all attributes as indicated in \cite{DPDN,APE}, still leading to a limited performance on challenging fine-grained datasets ({\it e.g.}, Flowers102\cite{FLO} for flower recognition and FGVC Aircraft\cite{Air} for airplane recognition).

\begin{figure}[t]
\centering
   \includegraphics[width=0.9\linewidth]{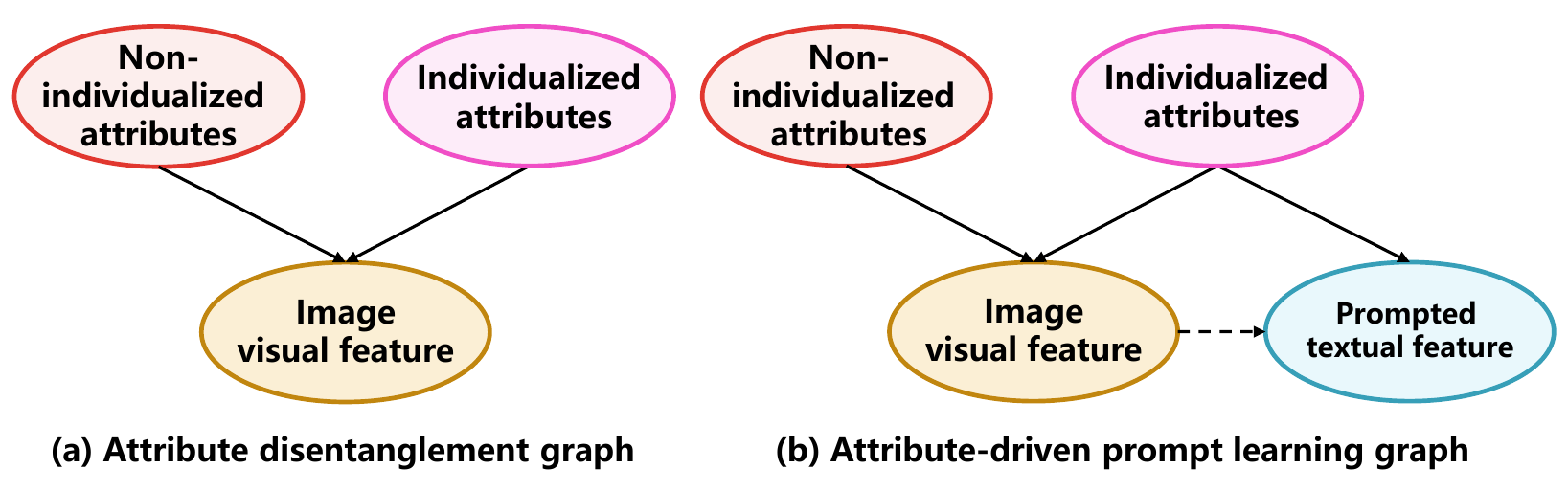}
   \caption{Causal graphs of (a) attribute disentanglement and (b) attribute-driven prompt learning for recognition.}
\label{fig:causal}
\end{figure}

\par Unlike the existing methods that either discard non-discriminative attributes or handle all attributes uniformly, our goal is to disentangle visual features into individualized attributes (exclusive to single classes) and non-individualized attributes (shared by some classes), and perceive all attributes according to their discrimination ability, assuming that all attributes contribute to prompt leaning but to varying degrees. This idea is supported by the following fact: In the fine-grained dataset Flowers102\cite{FLO}, petal color has weak discrimination ability, since more than 20 classes share yellow petals, yet it still aids in distinguishing classes with pink and yellow petals, whereas petal shape, such as trumpet-shaped or heart-shaped, is strongly discriminative as it varies across classes. To model these relationships, we introduce the attribute disentanglement graph shown in Fig. \ref{fig:causal}(a) to capture the connections among visual features and attributes. Based on this, we propose the attribute-driven prompt learning graph shown in Fig. \ref{fig:causal}(b)), which structures the relationship between a text prompt, visual features, and attributes. This graph could serve as a guidance for designing corresponding causal inference strategies that enable differentiated attribute perception for prompt learning.

\par Accordingly, we propose a causality-guided text prompt learning method, called CaPL, where attributes are disentangled from visual features to enable adaptive understanding through a visual granulation technique. Specifically, an attribute disentanglement module is firstly proposed to decompose the visual features extracted by CLIP\cite{CLIP} into non-individualized and individualized attribute representations, optimized within a Brownian Bridge Diffusion Model (BBDM)\cite{BBDM}-based network. Then, a granule learning module is proposed to construct sets of visual granules by integrating the disentangled attributes for recognition, where two causal inference strategies are applied: (1) For each of the individualized attributes, the factual intervention is to construct a corresponding factual granule by decorating this individualized attribute with all the non-individualized attributes; (2) To improve the generalizability of the text prompt, the counterfactual intervention is to construct counterfactual granules by swapping non-individualized and individualized attributes across different images. Finally, the text prompt is learned under the supervision of these visual granules during the training process.

\par Our contributions are summarized as follows: 
\begin{itemize}
    \item {We construct an attribute-driven prompt learning graph, which could depict the relationship between a text prompt, visual features and attributes. Accordingly, we propose an attribute disentanglement module to disentangle attributes with different discrimination ability;}
    \item {We explore the visual granulation technique in the proposed granule learning module under two causal inference strategies. The text prompt learned by utilizing the constructed visual granules as supervision signals could capture fine-grained discrepancies among different classes as demonstrated in Sec. \ref{sec:ablation};}
    \item {By integrating the above modules, the causality-guided text prompt learning method is proposed, whose superiority is demonstrated in Sec. \ref{sec:compare}.}
\end{itemize}

%% file: sec/2_relate.tex
\section{Related Work}
\label{sec:relate}

\par\noindent\textbf{Global prompt learning.} These methods learn prompts by utilizing features extracted by CLIP\cite{CLIP} holistically. Some of them project visual features to learn text prompts\cite{EMPL,APP, QCoOp,KAPT,LFA,ProMetaR,SHIP,KgCoOp,TCP,ProGrad}. Wang {\it et al.}\cite{SHIP} encoded visual features as a local bias of text prompts. Yao {\it et al.}\cite{TCP} incorporated prior knowledge into class-aware text prompts. Recently, some methods project textual features to learn visual prompts\cite{EVP,VPT,LoGoPrompt,ProVP,DPT}. Jia {\it et al.}\cite{VPT} introduced visual prompt before each image encoder layer. Shi {\it et al.}\cite{LoGoPrompt} synthesized class name images as visual prompts. More recently, some researchers use both text and visual prompts for dual-modal feature alignment\cite{TransAgent,COMMA,MaPLe,PromptSRC,HPT,KIM}. Khattak {\it et al.}\cite{MaPLe} injected dual-modal prompts into each CLIP encoder layer for mutual synergy. Hu {\it et al.}\cite{COMMA} regularized dual-modal prompts to retain pre-trained knowledge. However, these methods generally treat the visual and textual features as a whole, failing to capture differences between fine-grained classes and resulting in limited fine-grained recognition \cite{AAPE,CoCoLe}.

\par\noindent\textbf{Local prompt learning.} These methods extract attribute-specific features to capture subtle differences between fine-grained classes\cite{DEAL,APE,HPT,CDC,CoCoLe,CPL,GalLop}. Lafon {\it et al.}\cite{GalLop} selected top-k discriminative attributes for prompt learning. Wang {\it et al.}\cite{HPT} proposed hierarchical tuning to preserve only discriminative attributes in visual features. Zhang {\it et al.} extracted concept-level visual features describing attributes by a visual concept cache\cite{CPL} or a conceptual codebook\cite{CoCoLe}. Zhang {\it et al.}\cite{CDC} decoupled semantics from visual features to learn text prompts gradually. Despite their advancement in fine-grained recognition, they still fall short on challenging fine-grained datasets since they ignore that attributes contribute differently to recognition \cite{DPDN,APE}. Unlikely, we propose a causality-guided text prompt learning method via visual granulation, enabling perception of attributes based on their discrimination ability.

%% file: sec/3_method.tex
\section{Methodology}
\label{sec:method}

\subsection{Architecture}
\label{sec:architecture}

\begin{figure*}[t]
\centering
   \includegraphics[width=0.87\linewidth]{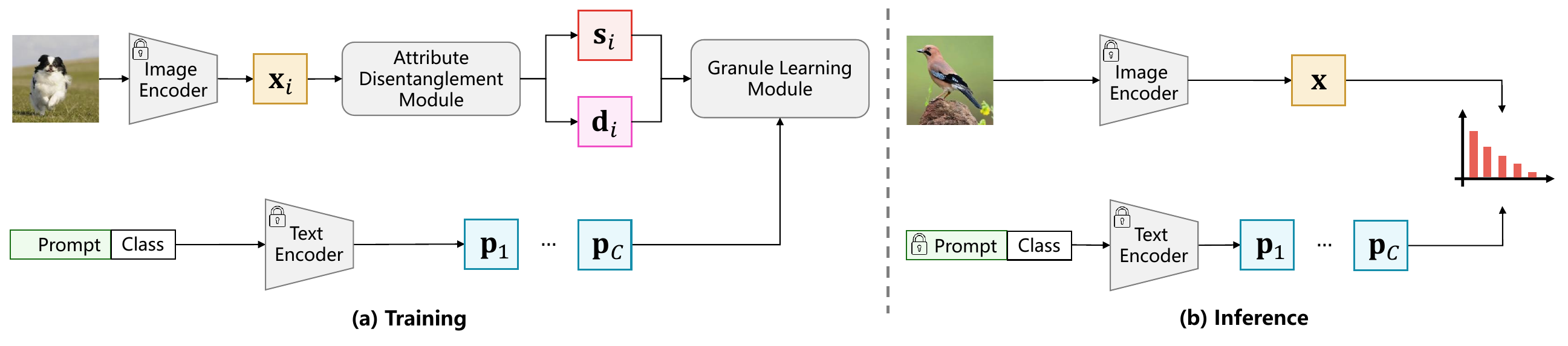}
   \caption{Architecture of CaPL, where (a) is the training stage and (b) is the inference stage. $\mathbf{x}_i$ and $\mathbf{x}$ are visual features, $\mathbf{s}_i$ and $\mathbf{d}_i$ are the non-individualized and individualized attribute representations, $\mathbf{p}_1,...,\mathbf{p}_C$ are the prompted textual features generated from a learnable text prompt and class names, $C$ is the number of classes, and the ``lock" symbol denotes the corresponding parameters are fixed.}
\label{fig:architecture}
\end{figure*}

\par Fig. \ref{fig:architecture} shows the architecture of the proposed causality-guided text prompt learning method, which contains a pre-trained CLIP consisting of an image encoder and a text encoder, an attribute disentanglement module, and a granule learning module to learn a text prompt.

\par As shown in Fig. \ref{fig:architecture}(a), at the training stage, the image encoder is used to extract visual features from images, and the text encoder is used to extract prompted textual features from a combination of a learnable text prompt and class names following\cite{KgCoOp,CoOp}. Then, the attribute disentanglement module decomposes the non-individualized and individualized attribute representations from visual features. According to the attribute-driven prompt learning graph in Fig. \ref{fig:causal}(b), the granule learning module uses both the non-individualized and individualized attributes disentangled from the attribute disentanglement module to construct visual granules, which are served as supervision signals for the text prompt to perceive specific attributes.

\par As shown in Fig. \ref{fig:architecture}(b), at the inference stage, a testing image is recognized by comparing the cosine similarity between its visual feature and all prompted textual features.

\subsection{Attribute Disentanglement Module}
\label{sec:CGM}

\begin{figure}[t]
\centering
   \includegraphics[width=0.87\linewidth]{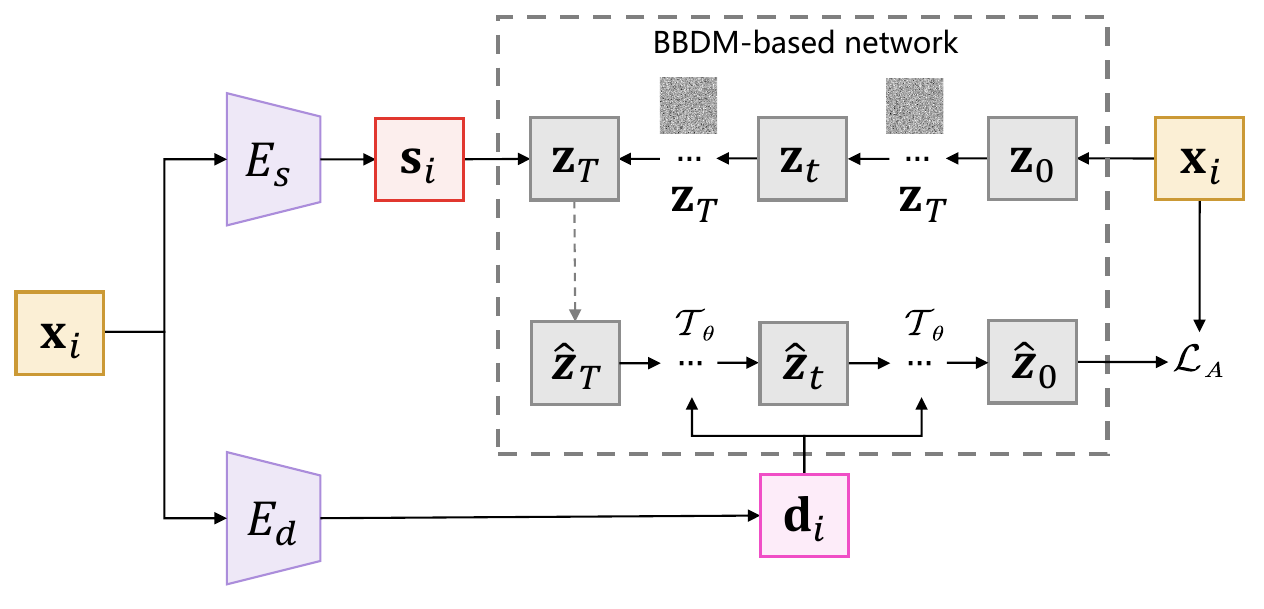}
   \caption{Architecture of attribute disentanglement module, which contains two encoders $E_s,E_d$ to extract non-individualized and individualized attribute representation $\mathbf{s}_i, \mathbf{d}_i$ from the visual feature $\mathbf{x}_i$ respectively, and a BBDM-based network. The upper feature transfer process of BBDM is the diffusion process, which generates latent features $\mathbf{z}_0,...,\mathbf{z}_T$. The lower one is the reverse process, which generate reconstructed features $\hat{\mathbf{z}}_T,...,\hat{\mathbf{z}}_0$ gradually, $\mathcal{T}_\theta$ is a learnable transfer model, and $\mathcal{L}_A$ is the training loss.}
\label{fig:ADM}
\end{figure}

\begin{figure*}[t]
\centering
   \includegraphics[width=0.77\linewidth]{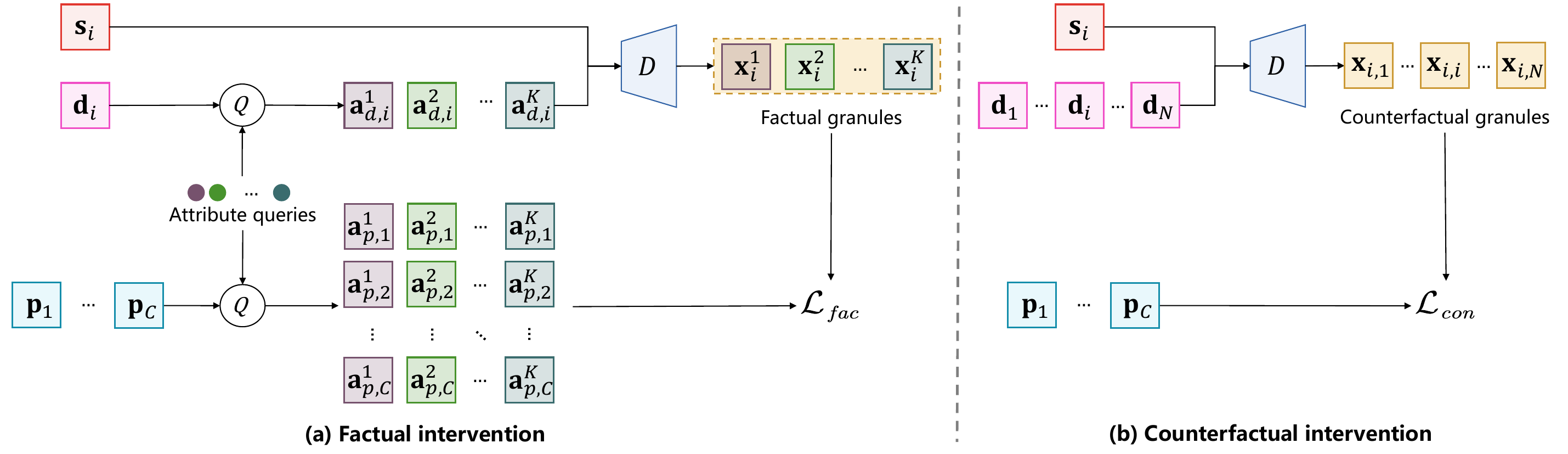}
   \caption{Architecture of granule learning module, which has two forms for (a) factual intervention and (b) counterfactual intervention. ``$Q$" is the query process, $\{\mathbf{a}_{d,i}^k\}_{k=1}^K$ and $\{\mathbf{a}_{p,c}^k\}_{c=1,k=1}^{C,K}$ are the visual and textual representations of each individualized attribute, $\{\mathbf{d}_i\}_{i=1}^N$ are $N$ individualized attributes in a training bath, $D$ is the decoder to generate visual granules, $\mathcal{L}_{fac},\mathcal{L}_{con}$ are training losses.}
\label{fig:GLM}
\end{figure*}

\par The attribute disentanglement module is proposed to decompose visual features into non-individualized and individualized attribute representations. In order to model the transition relationship among non-individualized attribute representations, individualized attribute representations, and the input visual features, a BBDM (Brownian Bridge Diffusion Model)\cite{BBDM} is introduced in this module as shown in Fig. \ref{fig:ADM}, inspired by the success of BBDM in modeling the relationship among different feature distributions in other tasks \cite{SBBDM,BBDM_Frame}, and ability of diffusion process to provide inductive biases for feature disentanglement\cite{DDPMdisentangle}.

\par Specifically, given the $i$-th image from class $c_i$, its visual feature $\mathbf{x}_i$ is extracted from the frozen CLIP image encoder. Two encoders $E_s$ and $E_d$ are used to obtain non-individualized attribute representation $\mathbf{s}_i$ and individualized attribute representation $\mathbf{d}_i$ respectively. 

\par Then, the BBDM is used to optimize the disentangled attributes by learning to reverse the disentanglement process. It is noted that the input visual features are regarded as target since they integrate both non-individualized and individualized attributes, while the individualized attribute representations serve as conditions due to their class-specific information, in contrast to the non-individualized ones that capture shared attributes. Therefore, we model the feature transition from non-individualized attribute representation $\mathbf{s}_i$ to the input visual feature $\mathbf{x}_i$ conditioned on the individualized attribute representation $\mathbf{d}_i$. Following \cite{BBDM}, the diffusion process transfers $\mathbf{z}_0=\mathbf{x}_i$ to $\mathbf{z}_T=\mathbf{s}_i$ following the Brownian Bridge stochastic process:

\vspace{-3mm}
\begin{equation}
    q\left(\mathbf{z}_{t}|\mathbf{z}_0, \mathbf{z}_T\right)=\mathcal{N}\left(\mathbf{z}_t;\left(1-m_t\right)\mathbf{z}_0+m_t\mathbf{z}_T,\delta_t\mathbf{I}\right)
\end{equation}

\par\noindent where $t\in[1,T]$, $T$ is the transfer step, $\mathbf{z}_t$ is a latent feature, $m_t=t/T$, $\delta_t=2(m_t-m_t^2)$, $\mathcal{N}(\mathbf{z};\mu,\Sigma)$ is a Gaussian distribution with mean $\mu$, variance $\Sigma$, $\mathbf{I}$ is an identity matrix. The reverse process learns to reverse the diffusion process by transferring $\hat{\mathbf{z}}_T=\mathbf{s}_i$ to $\hat{\mathbf{z}}_0=\hat{\mathbf{x}}_i$ conditioned on $\mathbf{d}_i$:

\begin{equation}
    p_\theta\left(\hat{\mathbf{z}}_{0:T}\right)=p\left(\hat{\mathbf{z}}_T\right)\prod_{t=1}^{T}p_\theta\left(\hat{\mathbf{z}}_{t-1}|\hat{\mathbf{z}}_t,\hat{\mathbf{z}}_T,\mathbf{d}_i\right)
\end{equation}

\par\noindent where $p_\theta\left(\hat{\mathbf{z}}_{t-1}|\hat{\mathbf{z}}_t,\hat{\mathbf{z}}_T,\mathbf{d}_i\right)$ follows a Gaussian distribution with mean $\mu_\theta\left(\hat{\mathbf{z}}_t,\hat{\mathbf{z}}_T,\mathbf{d}_i\right)$ and variance $\sigma^2_t=\frac{\delta_{t-1}}{\delta_t}\left[\delta_t-\delta_{t-1}\frac{\left(1-m_t\right)^2}{\left(1-m_{t-1}\right)^2}\right]$, $\hat{\mathbf{z}}_{t}$ is the reconstructed feature at time step $t$. The mean $\mu_\theta\left(\hat{\mathbf{z}}_t,\hat{\mathbf{z}}_T,\mathbf{d}_i\right)$ is parameterized as a learnable transfer model $\mathcal{T}_\theta\left(\hat{\mathbf{z}}_t,\hat{\mathbf{z}}_T,\mathbf{d}_i\right)$ to predict the difference $\Delta=\mathbf{z}_T-\mathbf{z}_0$. Following \cite{BBDM}, the transfer model $\mathcal{T}_\theta$ is trained by reconstructing $\mathbf{z}_0$ directly from a random $\mathbf{z}_t$ sampled from the diffusion process, rather than iterating $T$ steps following the reverse process. With $\mathbf{z}_t$, the difference $\hat{\Delta}_t=\mathcal{T}_\theta\left(\mathbf{z}_t,\mathbf{z}_T,\mathbf{d}_i\right)$ is predicted to obtain the restored $\Tilde{\mathbf{z}}_0=\mathbf{z}_T-\hat{\Delta}_t$, and the training loss $\mathcal{L}_A$ is to minimize the distance between restored $\Tilde{\mathbf{x}}_i=\Tilde{\mathbf{z}}_0$ and original $\mathbf{x}_i=\mathbf{z}_0$:

\begin{equation}
    \mathcal{L}_{A} = \parallel \Tilde{\mathbf{x}}_i-\mathbf{x}_i \parallel_2^2
\end{equation}

\par After training BBDM, the visual features could be effectively decomposed into two representations capturing non-individualized and individualized attributes respectively.

\subsection{Granule Learning Module}
\label{sec:FPM}

\par\noindent\textbf{Notation.} The text prompt is defined as $M$ learnable context vectors $\mathbf{v}=[\mathbf{v}_1,...,\mathbf{v}_M]$ following \cite{CoOp}. The prompted textual feature $\mathbf{p}_c$ of class $c$ is extracted by the fixed CLIP text encoder from a concatenation of $\mathbf{v}$ and $\mathbf{e}_c$, where $\mathbf{e}_c$ is the class name embedding obtained by CLIP tokenizer.

\par Then, the granule learning module is proposed to utilize the visual granulation technique for learning the text prompt. According to the attribute-driven prompt learning graph in Fig. \ref{fig:causal}(b), attributes contribute differently to learning a discriminative text prompt. To explicitly model these differences, the visual granulation technique constructs sets of visual granules as supervision signals by integrating different attributes under two causal inference strategies: 

\par\noindent\textbf{Factual intervention.} Since individualized attributes are highly discriminative while non-individualized attributes have relatively weak discrimination ability, the factual intervention is applied to construct factual granules, as shown in Fig. \ref{fig:GLM}(a), by decorating each of the individualized attributes with all non-individualized attributes to provide a complete yet focused representation for recognition. By serving as supervision signals for text prompt learning, the effect of single individualized attributes could be emphasized to enhance fine-grained recognition.

\par Specifically, a series of learnable attribute queries $\mathbf{q}^1,...,\mathbf{q}^K$, each corresponding to an individualized attribute, is defined to disentangle the visual representations $\{\mathbf{a}_{d,i}^k\}_{k=1}^K$ and textual representations $\{\mathbf{a}_{p,c}^k\}_{c=1,k=1}^{C,K}$ of each individualized attribute from $\mathbf{d}_i$ and $\mathbf{p}_1,...,\mathbf{p}_C$ respectively, where $K$ is a preset number of individualized attributes. For the $k$-th individualized attribute, its visual representation $\mathbf{a}_{d,i}^k$ is obtained as follows:

\begin{equation}
    \mathbf{a}_{d,i}^k=Q\left(\mathbf{d}_i,\mathbf{q}^k\right)=\mathcal{S}\left(\left(\mathbf{q}^k\right)^T\times \mathbf{d}_i/\sqrt{d}\right)\times \mathbf{d}_i
\end{equation}

\par\noindent where $Q$ is the query process, $\mathcal{S}(\cdot)$ is the softmax function, $(\cdot)^T$ is the transpose operation, and $d$ is the dimension of $\mathbf{d}_i$. The textual representation $\mathbf{a}_{p,c}^k$ of an arbitrary class $c$ can be obtained by: $\mathbf{a}_{p,c}^k=Q\left(\mathbf{p}_{c},\mathbf{q}^k\right)$ accordingly. 

\par Then, the factual granule $\mathbf{x}_i^k$ is constructed by combining $\mathbf{a}_{d,i}^k$ with the non-individualized representation $\mathbf{s}_i$: $\mathbf{x}_i^k=D\left(\mathbf{s}_i,\mathbf{a}_{d,i}^k\right)$, where $D$ is a MLP-based decoder. 

\par Finally, the textual representations are learned to recognize the factual granules. On the one hand, the textual representations $\{\mathbf{a}_{p,c}^k\}_{c=1,k=1}^{C,K}$ should recognize which individualized attribute constitutes the factual granule $\mathbf{x}_i^k$. The probability $p(y_a=k|\mathbf{x}_i^k)$ is defined as follows:

\begin{equation}
    p(y_a=k|\mathbf{x}_i^k)=\frac{{\rm exp}({\rm cos}(\mathbf{x}_i^k,\mathbf{a}_{p,c_i}^k)/\tau)}{\sum_{{k^\prime}=1}^K({\rm exp}({\rm cos}(\mathbf{x}_i^k,\mathbf{a}_{p,c_i}^{k^\prime})/\tau)}
\end{equation}

\par\noindent where $y_a$ is the predicted individualized attribute, $c_i$ is the class of $\mathbf{x}_i$, and $\tau$ is temperature coefficient. 

\par On the other hand,  the textual representations are learned to predict the class of $\mathbf{x}_i$ by calculating the cosine similarity between $K$ pairs of $\{\mathbf{x}_i^k\}_{k=1}^K$ and $\{\mathbf{a}_{p,c_i}^k\}_{k=1}^K$:

\begin{equation}
    p(y_v=c_i|\mathbf{x}_i)=\frac{\sum_{k=1}^K{\rm exp}({\rm cos}(\mathbf{x}_i^k,\mathbf{a}_{p,c_i}^k)/\tau)}{\sum_{c=1}^C\sum_{k=1}^K({\rm exp}({\rm cos}(\mathbf{x}_i^k,\mathbf{a}_{p,c}^k)/\tau)}
\end{equation}

\par\noindent where $y_v$ is the predicted class. The training loss $\mathcal{L}_{fac}$ is the weighted sum of two cross entropy losses:

\begin{equation}
\begin{adjustbox}{max width=0.9\linewidth}
$
\begin{aligned}
    \mathcal{L}_{fac} = -{\rm log}\left(p\left(y_a=k|\mathbf{x}_i^k\right)\right) -\lambda_f{\rm log}\left(p\left(y_v=c_i|\mathbf{x}_i\right)\right)
\end{aligned}
$
\end{adjustbox}
\end{equation}

\par\noindent where $\lambda_f$ is a constant weight. Note: The ablation study by omitting the non-individualized attributes is in Sec. \ref{sec:ablation}.

\par\noindent\textbf{Counterfactual intervention.} As stated in \cite{causal1,causal2}, homogeneous non-individualized attributes may lead to spurious correlations in recognizing visual features, which reduces the generalizability of the text prompt when encountering heterogeneous attributes. Therefore, the counterfactual intervention is applied to construct counterfactual granules, as shown in Fig. \ref{fig:GLM}(b), by mixing non-individualized and individualized attributes across images to simulate alternative contexts. By serving as another supervision signals, the generalizability of the text prompt is strengthened.

\par For the $i$-th image, given its $\mathbf{s}_i$ and all $\mathbf{d}_j,j=1,...,N$ in a training batch, where $N$ is batch size, the counterfactual granule $\mathbf{x}_{i,j}$ is obtained by: $\mathbf{x}_{i,j}=D\left(\mathbf{s}_i,\mathbf{d}_j\right)$. Then, the counterfactual granule $\mathbf{x}_{i,j}$ is assigned to the same class $c_j$ of the $j$-th image, from which $\mathbf{d}_j$ is decomposed. The probability $p\left(y_v=c_j|\mathbf{x}_{i,j}\right)$ for recognizing $\mathbf{x}_{i,j}$ is based on the cosine similarity between $\mathbf{x}_{i,j}$ and $\mathbf{p}_1,...,\mathbf{p}_C$.

\par Furthermore, to guarantee that the decoder $D$ could generate accurate visual granules, so that the recognition of $\mathbf{x}_{i,j}$ depends solely on the generalizability of the text prompt, we propose a reconstruction error to minimize the distance between $\mathbf{x}_{i,i}=D\left(\mathbf{s}_i,\mathbf{d}_i\right)$ and its corresponding input visual feature $\mathbf{x}_i$. The training loss $\mathcal{L}_{con}$ is the weighed sum of the cross entropy loss and reconstruction error:

\begin{equation}
    \mathcal{L}_{con}=-{\rm log}\left(p\left(y_v=c_j|\mathbf{x}_{i,j}\right)\right)+\lambda_r \parallel \mathbf{x}_i-\mathbf{x}_{i,i}\parallel _2^2
\end{equation}

\par\noindent where $\lambda_r$ is a constant weight. The total training objective $\mathcal{L}_G$ of the granule learning module is defined as:

\begin{equation}
    \mathcal{L}_{G} = \mathcal{L}_{fac} + \mathcal{L}_{con}
\end{equation}

\par It is noted that we do not apply the reconstruction error to factual granules, since they are constructed by utilizing only one individualized attribute and are actually different from their corresponding input visual features.

\subsection{Training and Inference}
\label{sec:objective}

\par The proposed CaPL utilizes an iterative training scheme, where each iteration contains two learning stages:

\par First, the attribute disentanglement module is learned by minimizing $\mathcal{L}_A$ in Eq. (3). Then, with fixed attribute disentanglement module, the text prompt and the granule learning module are learned by minimizing $\mathcal{L}_G$ in Eq. (9).

\par At the inference stage, the learned text prompt is used to generate prompted textual features of all classes. Then, the cosine similarity between a testing image visual feature and all prompted textual features is calculated for recognition.

%% file: sec/4_experiment.tex
\section{Experiments}
\label{sec:experiments}

\subsection{Tasks and Implementation Details}
\label{sec:setup}

\par\noindent\textbf{Tasks.} We evaluate the proposed CaPL in the following three downstream recognition tasks:

\par (1) \textbf{Base-to-new generalization}: As done in \cite{CoCoLe,CoCoOp}, our method is evaluated on 11 image recognition datasets listed in Table \ref{Tab.1}, and the classes in each dataset are equally divided into base and new classes. Our model is trained with 16-shot images on base classes, and tested on base and new classes. The evaluation metrics include the average per-class Top-1 accuracies on base classes ``Base" and new classes ``New", and the harmonic mean $H=\frac{2\times Base\times New}{Base+New}$.

\par (2) \textbf{Cross-dataset transfer}: As done in \cite{TCP,CoCoLe}, the 11 datasets used in the above task are also used here for testing the cross-dataset transfer ability of the proposed method. Our model is trained with 16-shot images from ImageNet-1K\cite{ImageNet} and then tested on the other 10 datasets. The evaluation metric is the average per-class Top-1 accuracy.

\par (3) \textbf{Cross-domain generalization}: As done in \cite{ProMetaR,CoCoLe}, our method is trained with 16-shot images from ImageNet-1K\cite{ImageNet}, and then tested on 4 variants listed in Table \ref{Tab.3} for verifying its cross-domain generalizability. The evaluation metric is the average per-class Top-1 accuracy.

\par\noindent\textbf{Implementation details.} The image and text encoders of CLIP are ViT-B-32\cite{ViT} and transformer\cite{Transformer}, which are pre-trained and kept frozen. The two encoders $E_s,E_d$ and decoder $D$ are two-layer MLPs. Following \cite{DDPM,BBDM}, the transfer model $\mathcal{T}_\theta$ is U-Net. The transfer step $T=1000$, the number of individualized attributes $K=10$, the batch size $N=64$, and the weights $\lambda_f,\lambda_r=1$. For the $n$-th iteration, the epochs of two learning stages are $10$ and $10n$.

\begin{table*}[!t]
    \centering
    \small
    \caption{Comparisons with the state-of-the-art CLIP-based prompt learning methods on base-to-new generalization on 11 public image recognition datasets. The best and second best results are marked in \textbf{bold} and \underline{underline.}}
    \vspace{-2mm}
    \label{Tab.1}
    \begin{subtable}{\textwidth}
    \centering
    \renewcommand{\arraystretch}{1.1}
    \scalebox{0.73}{
        \begin{tabular}{c|ccc|ccc|ccc|ccc}
            \hline
            \multirow{2}{*}{} & \multicolumn{3}{c|}{Average} & \multicolumn{3}{c|}{ImageNet-1K\cite{ImageNet}} & \multicolumn{3}{c|}{Caltech101\cite{Caltech101}} & \multicolumn{3}{c}{OxfordPets\cite{Pets}} \\
            & Base & New & H & Base & New & H & Base & New & H & Base & New & H \\
            \hline
            CLIP\cite{CLIP} & 69.34 & 74.22 & 71.70 & 72.43 & 68.14 & 70.22 & 96.84 & 94.00 & 95.40 & 91.17 & 97.26 & 94.12 \\
            CoOp\cite{CoOp} & 82.69 & 63.22 & 71.66 & 76.47 & 67.88 & 71.92 & 98.00 & 89.81 & 93.73 & 93.67 & 95.29 & 94.47 \\
            LoGoPrompt\cite{LoGoPrompt} & 84.47 & 74.24 & 79.03 & 76.74 & 70.83 & 73.66 & 98.19 & 93.78 & 95.93 & 96.07 & 96.31 & 96.18 \\
            COMMA\cite{COMMA} & 82.42 & 75.87 & 79.04 & 76.04 & 70.89 & 73.86 & 97.94 & 94.56 & 96.50 & 95.62 & 97.84 & 96.72 \\
            ProMetaR\cite{ProMetaR} & 84.39 & 76.93 & 80.49 & 77.76 & 70.75 & 74.09 & 98.11 & 94.29 & 96.16 & 95.57 & 97.43 & 96.49\\
            TCP\cite{TCP} & 84.13 & 75.36 & 79.51 & 77.27 & 69.87 & 73.38 & 98.23 & 94.67 & 96.42 & 94.67 & 97.20 & 95.92 \\
            HPT\cite{HPT} & 84.32 & 76.86 & 80.23 & 77.95 & 70.74 & 74.17 & \underline{98.37} & 94.98 & 96.65 & 95.78 & 97.75 & 96.71 \\
            CPL\cite{CPL} & 84.38 & 78.03 & 81.08 & 78.74 & 72.03 & 75.24 & 98.35 & 95.13 & 96.71 & 95.86 & 98.21 & 97.02 \\
            CoCoLe\cite{CoCoLe} & \underline{85.22} & \underline{80.31} & \underline{82.70} & \underline{79.25} & \underline{74.58} & \underline{76.84} & 98.17 & \underline{95.67} & \underline{96.90} & \underline{96.21} & \underline{98.55} & \underline{97.37} \\
            CDC\cite{CDC} & 83.34 & 77.38 & 80.25 & 77.50 & 71.73 & 74.51 & 98.20 & 94.37 & 96.25 & 96.07 & 98.00 & 97.02 \\
            \hline
            CaPL (ours) & \textbf{86.37} & \textbf{82.40} & \textbf{84.34} & \textbf{79.91} & \textbf{75.68} & \textbf{77.80} & \textbf{98.96} & \textbf{96.03} & \textbf{97.47} & \textbf{96.95} & \textbf{99.14} & \textbf{98.03} \\
            \hline
        \end{tabular}
        }
    \end{subtable}%
    \vspace{1mm}
    \begin{subtable}{\textwidth}
    \centering
    \renewcommand{\arraystretch}{1.1}
    \scalebox{0.73}{
    \begin{tabular}{c|ccc|ccc|ccc|ccc}
            \hline
            \multirow{2}{*}{} & \multicolumn{3}{c|}{StanfordCars\cite{Cars}} & \multicolumn{3}{c|}{Flowers102\cite{FLO}} & \multicolumn{3}{c|}{Food101\cite{Food}} & \multicolumn{3}{c}{FGVCAircraft\cite{Air}} \\
            & Base & New & H & Base & New & H & Base & New & H & Base & New & H \\
            \hline
            CLIP\cite{CLIP} & 63.37 & 74.89 & 68.65 & 72.08 & 77.80 & 74.83 & 90.10 & 91.22 & 90.66 & 27.19 & 36.29 & 31.09 \\
            CoOp\cite{CoOp} & 78.12 & 60.40 & 68.13 & 97.60 & 59.67 & 74.06 & 88.33 & 82.26 & 85.19 & 40.44 & 22.30 & 28.75 \\
            LoGoPrompt\cite{LoGoPrompt} & 78.36 & 72.39 & 75.26 & \textbf{99.05} & 76.52 & 86.34 & 90.82 & 91.41 & 91.11 & \textbf{45.98} & 34.67 & 39.53 \\
            COMMA\cite{COMMA} & 73.48 & 74.91 & 73.96 & 94.86 & 75.13 & 83.88 & 90.42 & 92.74 & 91.84 & 36.47 & 34.23 & 35.84 \\
            ProMetaR\cite{ProMetaR} & 78.32 & 75.18 & 76.72 & \underline{98.13} & 77.66 & 86.70 & 90.80 & 91.89 & 91.34 & 42.02 & 38.63 & 40.25 \\
            TCP\cite{TCP} & \underline{80.80} & 74.13 & 77.32 & 97.73 & 75.57 & 85.23 & 90.57 & 91.37 & 90.97 & 41.97 & 34.43 & 37.83 \\
            HPT\cite{HPT} & 76.95 & 74.23 & 75.57 & 98.17 & 78.37 & 87.16 & 90.46 & 91.57 & 91.01 & 42.68 & 38.13 & 40.28 \\
            CPL\cite{CPL} & 79.31 & 76.65 & 77.96 & 98.07 & 80.43 & 88.38 & 91.92 & 93.87 & 92.88 & 42.27 & 38.85 & 40.49 \\
            CoCoLe\cite{CoCoLe} & 80.32 & \underline{78.84} & \underline{79.57} & 97.72 & \underline{81.04} & \underline{88.60} & \underline{92.23} & \underline{94.28} & \underline{93.24} & 43.86 & \underline{42.65} & \underline{43.25} \\
            CDC\cite{CDC} & 73.80 & 73.97 & 73.88 & 96.93 & 75.07 & 84.61 & 90.87 & 92.33 & 91.59 & 37.47 & 37.50 & 37.48 \\
            \hline
            CaPL (ours) & \textbf{82.92} & \textbf{81.11} & \textbf{82.01} & 97.44 & \textbf{86.68} & \textbf{91.75} & \textbf{93.25} & \textbf{96.87} & \textbf{95.03} & \underline{45.85} & \textbf{44.81} & \textbf{45.32} \\
            \hline
        \end{tabular}
        }
    \end{subtable}%
    \vspace{1mm}
    \begin{subtable}{\textwidth}
    \centering
    \renewcommand{\arraystretch}{1.1}
    \scalebox{0.73}{
    \begin{tabular}{c|ccc|ccc|ccc|ccc}
            \hline
            \multirow{2}{*}{} & \multicolumn{3}{c|}{SUN397\cite{SUN}} & \multicolumn{3}{c|}{DTD\cite{DTD}} & \multicolumn{3}{c|}{EuroSAT\cite{EuroSAT}} & \multicolumn{3}{c}{UCF101\cite{UCF101}} \\
            & Base & New & H & Base & New & H & Base & New & H & Base & New & H \\
            \hline
            CLIP\cite{CLIP} & 69.36 & 75.35 & 72.23 & 53.24 & 59.90 & 56.37 & 56.48 & 64.05 & 60.03 & 70.53 & 77.50 & 73.85 \\
            CoOp\cite{CoOp} & 80.60 & 65.89 & 72.51 & 79.44 & 41.18 & 54.24 & 92.19 & 54.74 & 68.69 & 84.69 & 56.05 & 67.46 \\
            LoGoPrompt\cite{LoGoPrompt} & 81.20 & 78.12 & 79.63 & 82.87 & 60.14 & 69.70 & 93.67 & 69.44 & 79.75 & 86.19 & 73.07 & 79.09 \\
            COMMA\cite{COMMA} & 80.94 & 79.32 & 80.86 & 81.04 & 58.62 & 68.32 & 93.56 & 74.26 & 83.42 & 84.06 & 80.56 & 81.84 \\
            ProMetaR\cite{ProMetaR} & 82.70 & 79.02 & 80.82 & 83.02 & 64.05 & 72.31 & 94.94 & 77.44 & 85.30 & 86.97 & 79.84 & 83.25 \\
            TCP\cite{TCP} & 82.63 & 78.20 & 80.35 & 72.77 & 58.07 & 68.25 & 91.63 & 74.73 & 82.32 & 87.13 & 80.77 & 83.83 \\
            HPT\cite{HPT} & 82.57 & 79.26 & 80.88 & \underline{83.84} & 63.33 & 72.16 & 94.24 & 77.12 & 84.82 & 86.52 & 80.06 & 83.16 \\
            CPL\cite{CPL} & 81.88 & 79.65 & 80.75 & 80.92 & 62.27 & 70.38 & 94.18 & 81.05 & 87.12 & 86.73 & 80.17 & 83.32 \\
            CoCoLe\cite{CoCoLe} & \underline{83.97} & \underline{82.24} & \underline{83.10} & 82.46 & \underline{68.38} & \underline{74.76} & 95.03 & \underline{84.17} & \underline{89.27} & \underline{88.30} & \underline{83.05} & \underline{85.60} \\
            CDC\cite{CDC} & 82.37 & 80.03 & 81.18 & 82.70 & 64.10 & 72.22 & \underline{95.10} & 82.33 & 88.26 & 85.70 & 81.73 & 83.67 \\
            \hline
            CaPL (ours) & \textbf{84.66} & \textbf{83.03} & \textbf{83.84} & \textbf{84.56} & \textbf{71.23} & \textbf{77.32} & \textbf{95.96} & \textbf{85.97} & \textbf{90.69} & \textbf{89.60} & \textbf{85.81} & \textbf{87.66} \\
            \hline
        \end{tabular}
        }
    \end{subtable}
\end{table*}

\subsection{Comparative Evaluation}
\label{sec:compare}

\par The proposed CaPL is evaluated by comparing with 10 state-of-the-art prompt learning methods listed in Table \ref{Tab.1}. It is noted that CLIP\cite{CLIP} refers to directly use CLIP for evaluation without additional operation. The second to sixth methods are global prompt learning methods, while the last 4 are local methods. The results of the comparative methods are cited from their original papers. It is further noted that Tables \ref{Tab.2}-\ref{Tab.3} include some of the 10 comparative methods, as some were not evaluated on these tasks in their papers.

\par\noindent\textbf{Base-to-new generalization.} Table \ref{Tab.1} reports the results on base-to-new generalization task, where three points can be revealed: (1) For coarse-grained datasets ({\it e.g.}, \cite{Caltech101,UCF101,SUN}), both the global and local methods, as well as our proposed CaPL, achieve superior performance, demonstrating that the distinct differences between coarse-grained classes can be effectively captured; (2) For fine-grained datasets ({\it e.g.}, \cite{DTD,Cars,Air,FLO}), our CaPL significantly outperforms both the global and local comparative methods. For example, it improves the harmonic mean by 3.15$\%$ on Flowers102\cite{FLO}, 2.44$\%$ on StanfordCars\cite{Cars}, and 2.07$\%$ on FGVCAircraft\cite{Air}. These results indicate the effectiveness of the proposed visual granulation technique in capturing subtle discrepancies among fine-grained classes; (3) Across the average performance of all 11 datasets, our CaPL achieves superior recognition results, further indicating the effectiveness of our method for prompt learning.

\par\noindent\textbf{Cross-dataset transfer.} As shown in Table \ref{Tab.2}, which reports the results on cross-dataset transfer task, our CaPL achieves the best performance on the source dataset and 9 out of 10 target datasets. Notably, it delivers significant improvements on fine-grained datasets, such as 3.16$\%$ on Flowers102\cite{FLO} and 2.70$\%$ on StanfordCars\cite{Cars}. These results indicate the effectiveness of our CaPL in differentiated attribute perception via the visual granulation technique.

\begin{table*}[t]
\centering
\caption{Comparisons with comparative prompt learning methods on cross-dataset transfer. ImageNet-1K is used as source and the others are target. The datasets are denoted by abbreviations for simplicity. The best and second best results are marked in \textbf{bold} and \underline{underline}.}
\vspace{-2mm}
\scalebox{0.68}{
\begin{tabular}{ccccccccccccc}
\hline
\multirow{2}{*}{} & Source & \multicolumn{11}{c}{Target} \\ 
\cmidrule(r){2-2}\cmidrule(r){3-13}
 & ImageNet.\cite{ImageNet} & Caltech.\cite{Caltech101} & Pets.\cite{Pets} & Cars.\cite{Cars} & Flowers.\cite{FLO} & Food.\cite{Food} & Aircraft.\cite{Air} & SUN.\cite{SUN} & DTD\cite{DTD} & Euro.\cite{EuroSAT} & UCF.\cite{UCF101} & Avg.\\
\hline
CoOp\cite{CoOp} & 71.51 & 93.70 & 89.14 & 64.51 & 68.71 & 85.30 & 18.47 & 64.15 & 41.92 & 46.39 & 66.55 & 63.88 \\
COMMA\cite{COMMA} & 71.22 & 93.84 & 90.78 & 66.36 & 73.14 & 85.87 & 25.14 & 67.56 & 46.52 & 48.85 & 68.71 & 66.84 \\
TCP\cite{TCP} & 71.40 & 93.97 & 91.25 & 64.69 & 71.21 & 86.69 & 23.45 & 67.15 & 44.35 & 51.45 & 68.73 & 66.29 \\
HPT\cite{HPT} & 71.72 & 94.20 & \underline{92.63} & 66.33 & \underline{74.84} & 86.21 & 25.68 & \underline{68.75} & \underline{50.87} & 47.36 & 70.50 & 67.74 \\
CPL\cite{CPL} & 73.53 & 95.52 & 91.64 & 66.17 & 73.35 & 87.68 & 27.36 & 68.24 & 48.96 & 51.25 & 70.52 & 68.07 \\
CoCoLe\cite{CoCoLe} & \underline{73.88} & \textbf{95.88} & 91.93 & \underline{67.79} & 74.17 & \underline{87.97} & \underline{28.83} & \underline{68.75} & 49.26 & \underline{51.75} & \underline{72.78} & \underline{68.91} \\
CDC\cite{CDC} & 71.76 & 94.47 & 90.77 & 66.27 & 72.67 & 86.27 & 24.50 & 68.07 & 46.60 & 49.13 & 68.60 & 66.73 \\
\hline
CaPL (ours) & \textbf{75.02} & \underline{95.78} & \textbf{92.81} & \textbf{70.49} & \textbf{78.00} & \textbf{89.02} & \textbf{30.88} & \textbf{69.39} & \textbf{51.22} & \textbf{52.95} & \textbf{73.91} & \textbf{70.45} \\
\hline
\end{tabular}
}
\label{Tab.2}
\end{table*}

\begin{table}[t]
\centering
\caption{Comparisons with comparative prompt learning methods on cross-domain generalization. ImageNet-1K is used as source dataset and its 4 variants are used as target datasets. The best and second best results are marked in \textbf{bold} and \underline{underline}.}
\vspace{-2mm}
\scalebox{0.68}{
\begin{tabular}{ccccccc}
\hline
\multirow{2}{*}{} & Source & \multicolumn{4}{c}{Target} \\ 
\cmidrule(r){2-2}\cmidrule(r){3-7}
 & ImageNet.\cite{ImageNet} & -V2\cite{ImageNet-V2} & -S\cite{ImageNet-S} & -A\cite{ImageNet-A} & -R\cite{ImageNet-R} & Avg. \\
\hline
CoOp\cite{CoOp} & 71.51 & 64.20 & 47.99 & 49.71 & 75.21 & 59.28 \\
LoGoPrompt\cite{LoGoPrompt} & \textbf{75.27} & \textbf{66.65} & 48.99 & 51.36 & 76.85 & 60.96 \\
COMMA\cite{COMMA} & 71.22 & 64.84 & 49.65 & 51.64 & 77.56 & 60.92 \\
ProMetaR\cite{ProMetaR} & 71.29 & 64.39 & 49.55 & 51.25 & 77.89 & 60.77 \\
HPT\cite{HPT} & 71.72 & 65.25 & 49.36 & 50.85 & 77.38 & 60.71 \\
CPL\cite{CPL} & 73.53 & 65.18 & 49.92 & 50.73 & 77.38 & 60.80 \\
CoCoLe\cite{CoCoLe} & 73.88 & 65.86 & \underline{50.89} & \underline{51.75} & \underline{78.89} & \underline{61.85} \\
CDC\cite{CDC} & 71.76 & 64.87 & 50.33 & 50.40 & 78.10 & 60.93 \\
\hline
CaPL (ours) & \underline{75.02} & \underline{66.63} & \textbf{51.47} & \textbf{53.32} & \textbf{79.38} & \textbf{62.70} \\
\hline
\end{tabular}
}
\label{Tab.3}
\end{table}

\begin{table}[t]
\centering
\caption{Ablation study for different parts of CaPL.}
\vspace{-2mm}
\scalebox{0.73}{
\begin{tabular}{cccc}
\hline
Method & Base & New & H \\
\hline
CaPL w/o BBDM & 77.90 & 73.94 & 75.87 \\
CaPL w/o FAC & 77.32 & 73.58 & 75.40 \\
CaPL w/o CON & 78.79 & 72.32 & 75.42 \\
CaPL w/o Non-id & 78.58 & 74.59 & 76.53 \\
\hline
CaPL (ours) & \textbf{79.91} & \textbf{75.68} & \textbf{77.80} \\
\hline
\end{tabular}
}
\label{Tab.4}
\end{table}

\par\noindent\textbf{Cross-domain generalization.} Table \ref{Tab.3} reports the results on cross-domian generalization task. As is seen, our CaPL achieves the best performance on 3 variants of ImageNet, with the second best performance on ImageNet-V2\cite{ImageNet-V2}, demonstrating the generalizability of our proposed CaPL.

\subsection{Ablation Study and Model Analysis}
\label{sec:ablation}

\par In this subsection, all the experiments are conducted on ImageNet-1K\cite{ImageNet} via base-to-new generalization. 

\par\noindent\textbf{Ablation study.} We conduct ablation study on the BBDM-based network (denoted as ``BBDM"), factual intervention (denoted as ``FAC") and counterfactual intervention (denoted as ``CON"). We also evaluate the effectiveness of integrating non-individualized attributes in factual intervention (denoted as ``Non-id"). The corresponding results by omitting the four parts respectively are reported in Table \ref{Tab.4}. As is seen, all the four parts are effective in perceiving specific attributes for prompt learning, and the proposed CaPL achieves the best performance by utilizing all the four parts. 

\par\noindent\textbf{Analysis of attribute disentanglement module.} First, in Table \ref{Tab.5}, we evaluate two other optimization methods besides the BBDM-based network for attribute disentanglement: (1) Classification-based method that classifies the non-individualized and individualized attribute representations by maximizing the cross entropy loss for the non-individualized representations while minimizing it for individualized representations; (2) DDPM-based method that uses the denoising diffusion model (DDPM)\cite{DDPM}-based network to optimize the disentangled representations. Furthermore, we evaluate a variant of our BBDM-based method (denoted as ``BBDM-based variant") that regards the individualized representations as source while the non-individualized ones as condition. Four points can be revealed: (1) The classification-based method outperforms the method without additional optimization ({\it i.e.}, CaPL w/o BBDM in Table \ref{Tab.4}), showing the effectiveness of enforcing constraints on the disentangled representations; (2) The DDPM-based method performs worse than the two BBDM-based methods, since DDPM models the non-individualized representations as a noise distribution, which negatively impacts disentanglement accuracy; (3) The BBDM-based variant underperforms ours, mainly because the shared non-individualized representations are better suited as a common distribution for all visual features, while the class-specific individualized ones serve more effectively as conditions for guiding feature transitions to specific visual features; (4) The proposed BBDM-based network achieves the best performance, indicating the effectiveness of our proposed method to promote attribute disentanglement.

\begin{table}[t]
\centering
\caption{Ablation study for different optimization methods for
 attribute disentanglement.}
\vspace{-2mm}
\scalebox{0.73}{
\begin{tabular}{cccc}
\hline
Method & Base & New & H \\
\hline
Classification-based method & 78.44 & 74.24 & 76.28 \\
DDPM-based method & 79.12 & 74.93 & 76.97 \\
BBDM-based variant & 79.33 & 74.89 & 77.05 \\
\hline
BBDM-based method (ours) & \textbf{79.91} & \textbf{75.68} & \textbf{77.80} \\
\hline
\end{tabular}
}
\label{Tab.5}
\end{table}

\begin{figure}
    \centering
    \includegraphics[width=1\linewidth]{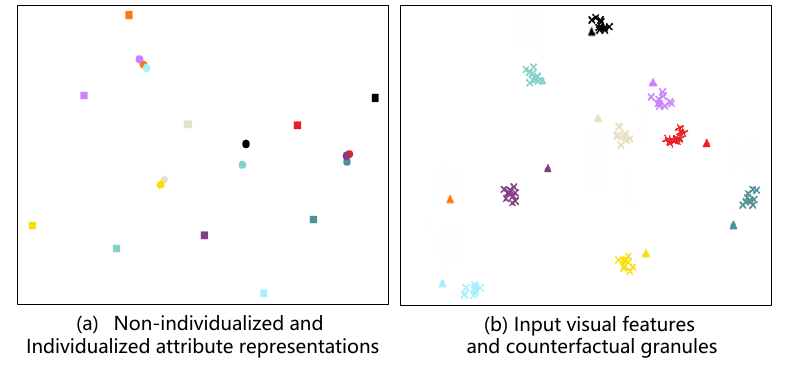}
    \caption{T-SNE visualizations of 10 images from 10 different classes of the StanfordCars dataset\cite{Cars}. (a) visualizes the non-individualized attribute representations (denoted as ``$\circ$") and individualized attribute representations (denoted as ``\scalebox{0.6}[1]{$\square$}"). (b) visualizes the input visual features (denoted as ``\scalebox{0.6}[1]{$\triangle$}") of the 10 images, and the counterfactual granules (denoted as ``\scalebox{0.7}[1]{$\times$}") constructed by swapping the attributes across the 10 images.}
    \label{fig:t-SNE}
\end{figure}

\begin{figure*}[t]
    \centering
    \includegraphics[width=1\linewidth]{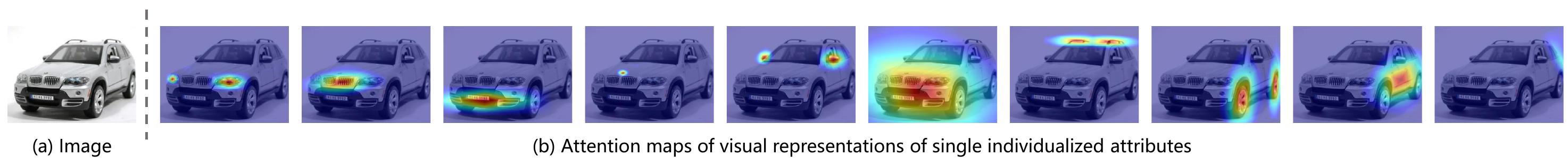}
    \caption{An image sample (a) and the attention maps (b) of its corresponding individualized attribute representations.}
    \label{fig:GRA-attention}
\end{figure*}

\par Then, we visualize the disentangled attribute representations in Fig. \ref{fig:t-SNE}(a), where the non-individualized representations (denoted as ``$\circ$") and individualized representations (denoted as ``\scalebox{0.6}[1]{$\square$}") disentangled from 10 images from the fine-grained StandfordCars dataset\cite{Cars} are visualized, where each image belongs to a different class. As is seen, the non-individualized representations are grouped into several small clusters, illustrating that these attributes have the same value across some classes but differ in other classes. In contrast, the individualized representations are scattered, further demonstrating that the attributes with different discrimination ability are well disentangled.

\par\noindent\textbf{Analysis of factual intervention.} First, we present visualizations to evaluate the effectiveness of the attribute queries in extracting representations of single individualized attributes. An image sample from the StanfordCars dataset\cite{Cars} is shown in Fig. \ref{fig:GRA-attention}(a), and its corresponding visual representations of all individualized attributes are visualized in Fig. \ref{fig:GRA-attention}(b) by utilizing the visualization method in \cite{APE} (Note: The number of individualized attributes is set to 10 in our experiments). As is seen, each attribute query could extract an individualized attribute-specific visual representation, focusing on a discriminative part of the target object. Furthermore, it is noted that the last visual representation does not contain clear information, likely because the corresponding individualized attribute relates to the car's tail, which is not visible in the image. This suggests that when an image lacks relevant information, the attribute query can extract a representation indicating the absence of clear information, further demonstrating its ability to extract attribute-specific representations.

\begin{table}[t]
\centering
\caption{Ablation study for attribute query initialization.}
\vspace{-2mm}
\scalebox{0.73}{
\begin{tabular}{cccc}
\hline
Method & Base & New & H \\
\hline
Non-learnable method & 77.83 & 72.75 & 75.20 \\
Prior initialization method & 78.34 & 74.12 & 76.17 \\
\hline
Random initialization method (ours) & \textbf{79.91} & \textbf{75.68} & \textbf{77.80} \\
\hline
\end{tabular}
}
\label{Tab.6}
\end{table}

\par Then, considering that the attribute queries are randomly initialized, we evaluate two other initialization methods in Table \ref{Tab.6}: (1) Non-learnable method that uses GPT-3\cite{GPT} to generate 10 phrases describing 10 individualized attributes that are discriminative for recognizing a class, and encodes these phrased by CLIP text encoder to act as fixed queries; (2) Prior initialization method that initializes the learnable attribute queries with the above queries obtained by GPT-3. The results in Table \ref{Tab.6} demonstrate the effectiveness of the proposed random initialization method in learning accurate attribute queries. It is noted that the prior initialization method underperforms the random method, mainly due to that random initialization could provide better coverage, improving representation and generalization in high-dimensional spaces as indicated in \cite{randominitialize}.

\begin{table}[t]
\centering
\caption{Ablation study for three hyperpatameters.}
\vspace{-2mm}
\scalebox{0.73}{
\begin{tabular}{c|ccccccc}
\hline
Value of $K$ & 0 & 2 & 4 & 6 & 8 & 10 & 12 \\
Accuracy (H) & 75.66 & 75.84 & 76.39 & 76.78 & 77.23 & \textbf{77.80} & 77.64 \\
\hline
Value of $\lambda_f$ & 0 & 0.2 & 0.4 & 0.6 & 0.8 & 1 & 1.2 \\
Accuracy (H) & 76.99 & 77.52 & 77.61 & 77.71 & 77.77 & \textbf{77.80} & 77.78 \\
\hline
Value of $\lambda_r$ & 0 & 0.2 & 0.4 & 0.6 & 0.8 & 1 & 1.2 \\
Accuracy (H) & 75.41 & 76.02 & 76.83 & 77.29 & 77.60 & \textbf{77.80} & 77.39 \\
\hline
\end{tabular}
}
\label{Tab.7}
\end{table}

\par\noindent\textbf{Analysis of counterfactual intervention.} We evaluate the effectiveness of constructing counterfactual granules in counterfactual intervention. Specifically, we select 10 images from 10 classes of StandfordCars\cite{Cars}, and construct the corresponding 100 counterfactual granules. Each granule is assigned the same class as the image from which its individualized attribute is disentangled. In Fig. \ref{fig:t-SNE}(b), we visualize the input visual features (denoted as ``\scalebox{0.6}[1]{$\triangle$}") and counterfactual granules (denoted as ``\scalebox{0.7}[1]{$\times$}"), with different colors indicating different classes. As shown, counterfactual granules that share the same individualized attribute ({\it i.e.}, same color) cluster around the corresponding visual features from which the individualized attribute is extracted, accurately reflecting the grouping of granules with the same class label near their real counterparts. This demonstrates the effectiveness of (1) the disentanglement of non-individualized and individualized attributes, (2) the decoder in generating counterfactual granules, and (3) the simulation of alternative contexts for improving generalization.

\par\noindent\textbf{Hyperparameters.} We analyze the number of individualized attributes $K$ and the weights $\lambda_f,\lambda_r$ in Table \ref{Tab.7}. Three points can be revealed: (1) Increasing $K$ improves accuracy, as disentangling more individualized attributes enhances finer perception. The performance stabilizes when $K>10$, indicating sufficient individualized attributes for recognition, so we set $K=10$; (2) The accuracy is insensitive to $\lambda_f$, so we set $\lambda_r=1$; (3) The best performance is achieved at $\lambda_r=1$, as it balances reconstruction error and cross entropy loss, so we set $\lambda_f=1$.

\par\noindent\textbf{Limitation.} Our CaPL has a relatively long training time due to the use of BBDM-based attribute disentanglement module. However, our method only uses the text prompt to calculate cosine similarity for inference, resulting in a faster inference time than some comparative methods while matching that of the others.

%% file: sec/5_conclusion.tex
\section{Conclusion}
\label{sec:conclusion}

\par In this paper, we construct an attribute-driven prompt learning graph for recognition, which depicts the relationship between a text prompt, visual features and attributes with different discrimination ability from a causal perspective. Accordingly, we propose a causality-guided text prompt learning method, where a visual granulation technique could capture subtle discrepancies among fine-grained classes by constructing sets of visual granules as supervision by integrating attributes under two causal inference strategies. Experimental results have demonstrated the superiority of the proposed method. In future, we would investigate how to construct more accurate granules for prompt learning.
\par\noindent\textbf{Acknowledgements.} This work was supported by the National Natural Science Foundation of China (Grant Nos. 62376269, 61991423, U1805264), the Beijing Municipal Science and Technology Project (Grant No. Z211100011021004).